\documentclass[letterpaper,10pt,conference]{ieeeconf}
\IEEEoverridecommandlockouts
\usepackage{amsmath,amssymb,amsfonts}
\usepackage{graphicx,microtype,booktabs,multirow,bm}
\makeatletter
\let\NAT@parse\undefined
\makeatother
\usepackage[numbers,sort&compress]{natbib}
\usepackage{xcolor,dblfloatfix}
\newcommand{\SO}{\mathrm{SO}}
\newcommand{\SE}{\mathrm{SE}}

\newcommand{\diag}{\mathrm{diag}}

\newcommand{\I}{\mathbf I}
\newcommand{\0}{\mathbf 0}
\newcommand{\R}{\mathbf R}
\newcommand{\p}{\mathbf p}
\renewcommand{\v}{\mathbf v}
\newcommand{\g}{\mathbf g}
\newcommand{\bg}{\mathbf b_{g}}
\newcommand{\ba}{\mathbf b_{a}}

\newcommand{\dbg}{\dot{\mathbf b}_{g}}
\newcommand{\dba}{\dot{\mathbf b}_{a}}
\newcommand{\bx}{\mathbf x}
\newcommand{\bX}{\mathbf X}
\newcommand{\bP}{\mathbf P}

\newcommand{\bH}{\mathbf H}
\newcommand{\bK}{\mathbf K}

\newcommand{\bPhi}{\boldsymbol\Phi}
\newcommand{\dt}{\Delta t}
\begin{document}

\title{\LARGE\bfseries SCF-NIO: Temporal Neural Inertial Odometry in a Shared Canonical Frame}

\author{Qizhi Guo, Defu Lin, Shaoming He%
\thanks{This work was supported by the National Natural Science Foundation of China under Grant 52302449.}%
\thanks{Authors' addresses: Qizhi Guo, Defu Lin, and Shaoming He are with the School of Aerospace Engineering and the Beijing Key Laboratory of UAV Autonomous Control, Beijing Institute of Technology, Beijing 100081, China, E-mail: ({qizhi.guo@bit.edu.cn}; {lindf@bit.edu.cn}; {shaoming.he@bit.edu.cn}). \textit{(Corresponding author: Shaoming He.)}}%
}

\maketitle
\thispagestyle{empty}
\pagestyle{empty} 

\begin{abstract}

Inertial measurement units (IMUs) offer low cost motion tracking, yet pure inertial integration suffers from rapid trajectory drift. Neural inertial odometry reduces this drift with learned displacement priors. However, inconsistent motion representations can limit temporal displacement estimation, reducing the quality of the resulting priors. To address this, we present Shared Canonical Frame Neural Inertial Odometry (SCF-NIO), which incorporates temporal displacement priors into inertial state estimation. An equivariant frame network expresses IMU measurement sequences in a shared canonical frame, providing a consistent representation across windows. A temporal network uses this representation to estimate displacement and its uncertainty. The resulting displacement priors are fused with inertial propagation in an error state Kalman filter. Evaluations on TLIO and OxIOD show lower mean absolute trajectory error than three neural inertial baselines. Relative to TLIO, SCF-NIO reduces this error by 22.9\% and 17.5\% on the respective datasets.

\end{abstract}


\section{Introduction}
\label{sec:introduction}

Mobile robots and wearable devices need continuous estimates of their own motion for navigation and tracking.
Visual inertial odometry (VIO) is the efficient solution, correcting the integration of inertial measurement unit (IMU) data with camera observations~\cite{qin2018vinsmono,campos2021orbslam3}.
Its visual front end relies on feature extraction or optical flow tracking across images.
This tracking fails under extreme lighting or in weakly textured scenes~\cite{zhao2021super}.
The estimator then integrates biased inertial measurements without correction, and its trajectory diverges within seconds~\cite{buchanan2023deep}.

IMUs measure motion regardless of illumination and scene texture, which has drawn growing attention to inertial odometry from an IMU alone.
Because direct integration drifts, recent methods learn motion patterns from IMU data instead~\cite{herath2020ronin}.
In the formulation of TLIO, a convolutional neural network (CNN) maps a one second window of gravity aligned measurements to displacement and its covariance, and an extended Kalman filter (EKF) fuses this neural displacement prior with inertial propagation~\cite{liu2020tlio}.
This formulation has been applied to legged robots and drones~\cite{buchanan2022learning,cioffi2023learned}, while pedestrian applications approach the accuracy of VIO~\cite{cioffi2023learned}.
These priors help constrain inertial drift when reliable visual observations are unavailable.

\begin{figure}[!t]
\centering
\includegraphics[width=0.95\columnwidth]{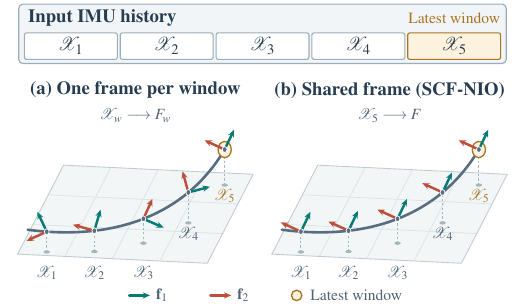}
\caption{Canonical frames for the same IMU history. (a) Each window $\mathcal X_w$ defines its own frame $F_w$. (b) SCF-NIO constructs $F$ from the latest window $\mathcal X_5$ and expresses all five windows in this frame.}
\label{fig:symmetry}
\end{figure}

The representation of the IMU data limits these priors.
Gravity alignment leaves the heading arbitrary, so one motion appears in infinitely many coordinate representations.
Random yaw augmentation and consistency losses~\cite{cao2022rio} encourage motion predictions to rotate with the input, yielding approximate yaw equivariance~\cite{jayanth2025lieevents}.
The predicted covariance must also transform consistently under the same rotation.
EqNIO removes this variation by expressing each window in a learned canonical frame~\cite{jayanth2025eqnio}.
This frame is predicted from a single one second window and is rebuilt for every window.

A short input window also limits the temporal context of these priors.
Longer windows lower the error of each prediction, but their overlap correlates consecutive errors, and the trajectory error of TLIO barely changes~\cite{liu2020tlio}.
Recurrent models such as long short term memory (LSTM) networks exploit longer IMU histories but rely on rotation augmentation~\cite{chen2021rnin}.
Canonical frames replace this augmentation, but EqNIO builds a new frame for each window, so a recurrent model confuses the frame changes with motion, as Fig.~\ref{fig:symmetry} illustrates.
We present SCF-NIO, a temporal neural inertial odometry built on a shared canonical frame.
It expresses the IMU history in one shared canonical frame, so that a temporal network learns motion patterns in a consistent representation.
Our contributions are:
\begin{enumerate}
\item We propose SCF-NIO, a neural inertial odometry method that expresses successive IMU windows in a shared canonical frame. The latest window defines the horizontal axes for the full history, preserving consistent motion geometry across windows.
\item We learn a temporal displacement prior by combining features from successive windows in the shared frame. The network estimates displacement and covariance for the latest window, providing a relative motion constraint for inertial state estimation.
\item Extensive experiments on TLIO and OxIOD datasets demonstrate lower mean absolute trajectory error than three neural inertial baselines, complemented by ablations validating the shared frame and temporal context.
\end{enumerate}

\section{Related Work}
\label{sec:related_work}

\subsection{Neural Inertial Odometry}
\label{subsec:rw_learning}

RoNIN regresses planar velocity in a gravity aligned frame defined by the device orientation, whose errors distort its integrated trajectory~\cite{herath2020ronin}.
IONet integrates planar displacement regressed from IMU windows without any uncertainty, so its position error accumulates uncorrected~\cite{chen2018ionet}.
TLIO instead fuses the 3D displacement over a one second window and its covariance in a stochastic cloning EKF. The covariance is diagonal in a gravity aligned frame with the device heading, so its principal axes follow the device rather than the motion~\cite{liu2020tlio}.
Its adaptations to legged robots and VIO keep this covariance~\cite{buchanan2022learning,chen2021rnin}, and its drone racing variant needs collective thrust and uses a constant measurement noise~\cite{cioffi2023learned}.
AI-IMU and RINS-W depend on pseudo measurements such as null lateral velocity, which hold only for wheeled vehicles~\cite{brossard2020aiimu,brossard2019rins}.
Random rotations about gravity in training leave TLIO and RoNIN unconstrained~\cite{liu2020tlio,herath2020ronin}, and TLIO still estimates different trajectories for rotated inputs~\cite{jayanth2025eqnio}.
RIO penalizes inconsistent predictions on rotated inputs but drops this loss below 0.5~m/s, where the direction of motion is ambiguous~\cite{cao2022rio}.
AirIO pairs body frame data of a drone with its full attitude and thus reintroduces the heading~\cite{qiu2025airimu}.
LieEvent-NIO~\cite{jayanth2025lieevents} converts IMU data into events on $\SE(3)$, which reduce variation in speed and sampling rate but not in heading.

\subsection{Canonical Frames for Temporal Modeling}
\label{subsec:rw_equivariance}

Gravity alignment leaves the horizontal axes of an IMU sequence undetermined.
RoNIN~\cite{herath2020ronin} keeps a fixed but arbitrary heading reference for each sequence and relies on yaw augmentation to learn how velocity transforms under a change of reference.
CTIN~\cite{rao2022ctin} applies temporal attention in a navigation frame whose horizontal axes depend on the initial device heading.
Learned canonicalization defines reference axes from the input itself~\cite{kaba2023equivariance}.
EqNIO~\cite{jayanth2025eqnio} predicts canonical axes separately for each input window, so the same motion direction can have different coordinates in successive windows.
The available motion history also depends on how the predictor processes successive windows.
IONet~\cite{chen2018ionet} estimates distance and heading change independently for each window.
The predictor therefore infers initial velocity without access to motion before the window.
IDOL~\cite{sun2021idol} limits drift at inference by resetting its displacement predictor's hidden state between windows, discarding the motion history encoded in that state.
X-IONet~\cite{shen2026xionet} models temporal and cross-axis dependencies with dual-stage Transformer attention, but its displacement predictor sees only one window.
RNIN-VIO~\cite{chen2021rnin} extends recurrence across ten one-second windows, relying on random yaw rotations during training to learn how displacement changes with the heading reference.
Tartan IMU~\cite{zhao2025tartan} estimates velocity from ten windows of body frame measurements, so successive predictions refer to different axes as the platform turns.

\section{Preliminaries and Problem Formulation}
\label{sec:problem_statement}

Consider an IMU rigidly attached on a moving platform.
The IMU defines the body frame, and the $z$ axis of the fixed world frame points opposite to gravity $\g=[0\;0\;{-g}]^T$, where $g>0$ is the gravitational acceleration.
The gyroscope and accelerometer measurements are modeled as
\begin{equation}
 \tilde{\boldsymbol\omega}=\boldsymbol\omega+\bg+\mathbf n_g,\qquad
 \tilde{\mathbf a}=\R^T(\mathbf a-\g)+\ba+\mathbf n_a
 \label{eq:imu}
\end{equation}
where $\R\in\SO(3)$ rotates vectors from the body frame to the world frame, $\boldsymbol\omega$ is the angular velocity in the body frame, and $\mathbf a$ is the acceleration in the world frame.
The biases $\bg$ and $\ba$ are expressed in the body frame, and $\mathbf n_g$ and $\mathbf n_a$ are zero mean white Gaussian noises.
The navigation state consists of orientation, velocity, position, and IMU biases
\begin{equation}
 \bx=(\R,\v,\p,\bg,\ba)
 \label{eq:state}
\end{equation}
where $\v,\p\in\mathbb R^3$ are the velocity and position in the world frame.
From~\eqref{eq:imu}, the state evolves as
\begin{equation}
 \begin{aligned}
 \dot\R&=\R\,[\tilde{\boldsymbol\omega}-\bg-\mathbf n_g]_\times&
 \dot\v&=\R(\tilde{\mathbf a}-\ba-\mathbf n_a)+\g\\
 \dot\p&=\v \quad \dbg=\boldsymbol\eta_g&
 \dba&=\boldsymbol\eta_a
 \end{aligned}
 \label{eq:kinematics}
\end{equation}
where $[\cdot]_\times$ denotes the skew symmetric matrix, and the biases follow random walks driven by the white noises $\boldsymbol\eta_g$ and $\boldsymbol\eta_a$.
Integrating~\eqref{eq:kinematics} with the noisy $\tilde{\boldsymbol\omega}$ and $\tilde{\mathbf a}$ accumulates drift.
A learned prior therefore constrains the displacement between the retained pose at $t_i$ and the current pose at $t_j$.
We express this displacement in the gravity aligned frame at $t_i$ as
\begin{equation}
 \mathbf d=\R_\gamma^T(\p_j-\p_i)
 \label{eq:displacement}
\end{equation}
where $\R_\gamma$ is the yaw rotation of $\R_i$.
This frame removes the unobservable global yaw from $\mathbf d$.
We denote the IMU measurements in~\eqref{eq:imu} expressed in the gravity aligned frame at $t_i$ by $\tilde{\boldsymbol\omega}^{g}$ and $\tilde{\mathbf a}^{g}$. These samples form the history $\mathcal X=(\mathcal X_1,\ldots,\mathcal X_W)$ of $W$ consecutive windows, each with $L$ samples, and the latest window $\mathcal X_W$ spans $[t_i,t_j]$.
The IMU does not measure $\mathbf d$ directly, so a neural network learns it from $\mathcal X$.
The network $f$ regresses a displacement estimate $\hat{\mathbf d}$ and its covariance $\boldsymbol\Sigma$, which parameterize the neural displacement prior used in the learned measurement model
\begin{equation}
 \mathbf d=\hat{\mathbf d}+\mathbf n_d,\qquad
 (\hat{\mathbf d},\boldsymbol\Sigma)=f(\mathcal X)
 \label{eq:measurement}
\end{equation}
with $\mathbf n_d\sim\mathcal N(\mathbf 0,\boldsymbol\Sigma)$.
The network output $\hat{\mathbf d}$ therefore acts as a displacement measurement, and $\boldsymbol\Sigma$ sets its weight in the filter update.
This relative measurement couples the current pose to the retained pose at $t_i$.

\noindent\textbf{Problem 1.} Given the initial state $\bx_0$ and the measurements $\tilde{\boldsymbol\omega}$ and $\tilde{\mathbf a}$ up to $t_j$, learn the network $f$ that regresses $(\hat{\mathbf d},\boldsymbol\Sigma)$ from $\mathcal X$, and then estimate the pose at $t_j$ as
\begin{equation}
 (\hat\R_j,\hat\p_j)=\mathcal E\bigl(\bx_0,\tilde{\boldsymbol\omega},\tilde{\mathbf a};f\bigr)
 \label{eq:odometry}
\end{equation}
where the filter $\mathcal E$ propagates the state by~\eqref{eq:kinematics} and updates it with the learned measurement~\eqref{eq:measurement}.

\section{Methodology}
\label{sec:methodology}

\begin{figure}[!t]
\centering
\includegraphics[width=\columnwidth,trim=0 10bp 0 5bp,clip]{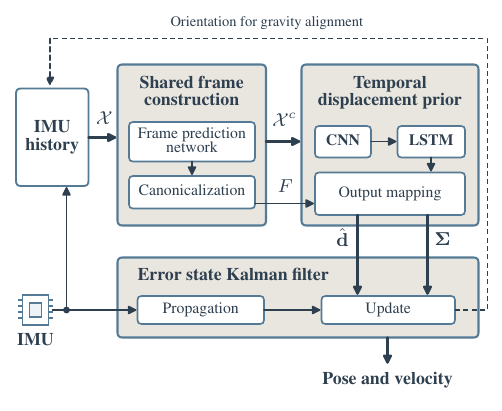}
\caption{Overview of SCF-NIO. The frame prediction network predicts one canonical frame from the latest window, and this frame canonicalizes every window of the gravity aligned IMU history. A CNN and an LSTM regress the displacement and its covariance in canonical coordinates, which are mapped back to the gravity aligned frame as the filter measurement.}
\label{fig:pipeline}
\end{figure}

SCF-NIO combines a neural displacement prior with an error state Kalman filter (ESKF), as shown in Fig.~\ref{fig:pipeline}. The network $f$ predicts the prior $(\hat{\mathbf d},\boldsymbol\Sigma)$ over $[t_i,t_j]$ from the IMU history $\mathcal X$.
The ESKF incorporates this prior through the measurement model~\eqref{eq:measurement}.

\subsection{Shared Frame Construction}
\label{subsec:frame}
% logic: build F, share F

For each IMU history $\mathcal X$,
we construct one canonical frame $F$ shared by all $W$ windows
\begin{equation}
 F=\diag([\mathbf f_1\ \mathbf f_2],1)
 \label{eq:canonicalization}
\end{equation}
where $\mathbf f_1,\mathbf f_2\in\mathbb R^2$ form the orthonormal horizontal basis.
\begin{figure}[!h]
\centering
\includegraphics[width=1.00\columnwidth]{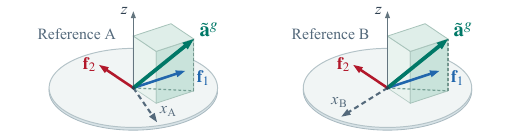}
\caption{Schematic of the measured acceleration $\tilde{\mathbf a}^{g}$ under two heading references. The dashed arrows mark the reference axes $x_A$ and $x_B$. In both views, the acceleration has the same components along the horizontal canonical axes $\mathbf f_1$ (blue) and $\mathbf f_2$ (red) and the common vertical $z$ axis.}
\label{fig:canonical-coordinates}
\end{figure}

Canonicalization makes the IMU representation independent of the heading reference, and the shared axes preserve relative heading changes between windows, as shown in Fig.~\ref{fig:canonical-coordinates}.

To construct this frame, we propose a shared frame network.
Within the latest IMU window $\mathcal X_W$, the network uses convolution across neighboring samples to extract horizontal vector features.
The network averages these features to predict two horizontal vectors $\mathbf r_1,\mathbf r_2\in\mathbb R^2$.
Applying the Gram--Schmidt process to these vectors gives the horizontal axes of $F$, $\mathbf f_1=\mathbf r_1/\|\mathbf r_1\|$ and $\mathbf f_2=\mathbf r_2^\perp/\|\mathbf r_2^\perp\|$.
Here $\mathbf r_2^\perp=\mathbf r_2-(\mathbf f_1^T\mathbf r_2)\mathbf f_1$ is the component of $\mathbf r_2$ perpendicular to the first axis $\mathbf f_1$.
If either predicted vector is near zero or the two are nearly collinear, we search the IMU history for a horizontal reference direction.
We express each accelerometer and gyroscope sample in $F$ as
\begin{equation}
 \tilde{\mathbf a}^{c}=F^T\tilde{\mathbf a}^{g},\qquad
 \tilde{\boldsymbol\omega}^{c}=\det(F)F^T\tilde{\boldsymbol\omega}^{g}
 \label{eq:imu_group_action}
\end{equation}
where the superscript $c$ denotes coordinates in $F$, and $\det(F)$ accounts for the sign change of an axial vector under reflection.
Under a change of input coordinates, the frame in~\eqref{eq:canonicalization} and the canonicalized history, viewed as functions of $\mathcal X$, satisfy
\begin{equation}
 \begin{aligned}
 F(Q\cdot\mathcal X)&=QF(\mathcal X)\\
 \mathcal X^c(Q\cdot\mathcal X)&=\mathcal X^c(\mathcal X)
 \end{aligned}
 \label{eq:invariance}
\end{equation}
where $Q$ belongs to the group $\mathcal G$ of gravity preserving rotations and reflections, $Q\cdot\mathcal X$ applies $Q$ to each sample as in~\eqref{eq:imu_group_action}, and $\mathcal X^c=(\mathcal X_1^c,\ldots,\mathcal X_W^c)$ is the history of canonicalized samples.
Changing the heading reference rotates both the IMU data and the predicted axes, so each window retains the same coordinates in the shared frame.

\subsection{Temporal Prior in a Shared Frame}
\label{subsec:temporal}

The temporal network estimates the displacement and its covariance over the latest window from the canonicalized IMU history.
The shared frame maps these predictions to the gravity aligned frame to form the neural displacement prior
\begin{equation}
 \hat{\mathbf d}=F\hat{\mathbf d}^c,\qquad
 \boldsymbol\Sigma=F\diag\bigl(e^{2\mathbf u}\bigr)F^T
 \label{eq:decanonicalization}
\end{equation}
The canonical outputs $\hat{\mathbf d}^c$ and $\mathbf u$ denote the displacement estimate and the log standard deviations, respectively.
The diagonal covariance $\diag(e^{2\mathbf u})$ is expressed in the same canonical frame.
Using the same axes throughout the IMU history lets the temporal network compare motion across successive windows.
The network regresses the canonical displacement and uncertainty from a common temporal representation
\begin{equation}
 (\hat{\mathbf d}^c,\mathbf u)=\psi(\mathbf h_W)
 \label{eq:canonical_cov}
\end{equation}
where $\psi$ denotes two fully connected output branches.
Both branches use $\mathbf h_W$, the final LSTM hidden state summarizing the $W$ canonicalized windows.
The LSTM incorporates motion context from earlier windows by aggregating canonicalized window features in temporal order
\begin{equation}
 \mathbf h_w=\mathrm{LSTM}\bigl(\mathbf h_{w-1},\phi(\mathcal X^c_w)\bigr),\quad w=1,\dots,W
 \label{eq:lstm}
\end{equation}
where $\mathcal X^c_w$ is the $w$-th window of $\mathcal X^c$ and $\phi$ is a CNN encoder with weights shared across windows.
The hidden state $\mathbf h_w$ summarizes the first $w$ windows.
We initialize $\mathbf h_0=\0$ for each history because each history uses its own shared frame.
Since $\mathcal X^c$ in~\eqref{eq:invariance} is invariant under $\mathcal G$, both the recurrent state $\mathbf h_w$ in~\eqref{eq:lstm} and the canonical outputs in~\eqref{eq:canonical_cov} remain invariant under $\mathcal G$.
Mapping these outputs through the equivariant frame $F$ in~\eqref{eq:decanonicalization} therefore yields a temporal displacement prior $(\hat{\mathbf d},\boldsymbol\Sigma)$ that captures multi window motion context while transforming consistently with the input history.

\subsection{Training and Filter Integration}
\label{subsec:integration}

We jointly train all network components from random initialization using displacement supervision
\begin{equation}
 \mathcal L=\tfrac16\bigl[(\mathbf d-\hat{\mathbf d})^T\boldsymbol\Sigma^{-1}(\mathbf d-\hat{\mathbf d})+\log\det\boldsymbol\Sigma\bigr]
 \label{eq:nll}
\end{equation}
where $\mathbf d$ is the reference displacement over the latest window and $\mathcal L$ is the Gaussian negative log likelihood (NLL), averaged over three axes with the constant omitted.
Gradients of $\mathcal L$ propagate through the decanonicalization in~\eqref{eq:decanonicalization} and the canonicalization in~\eqref{eq:imu_group_action} to the shared frame network, so the frame is learned without frame labels.
Since the NLL penalizes both the displacement error and a mismatched $\boldsymbol\Sigma$, the predicted covariance reflects the displacement uncertainty and can weight the displacement prior in a filter.
The ESKF fuses this prior through the measurement model~\eqref{eq:measurement}, with $\boldsymbol\Sigma$ as the measurement noise covariance.
The filter propagates $\bx$ and the covariance
\begin{equation}
 \bP_{k+1}^{-}=\bPhi_k\bP_k^{+}\bPhi_k^T+\mathbf B_k\mathbf N_k\mathbf B_k^T
 \label{eq:propagation}
\end{equation}
where $\bP$ is the error covariance, $k$ indexes IMU samples, and superscripts $-$ and $+$ denote prior and posterior quantities.
The matrices $\bPhi_k$, $\mathbf B_k$, and $\mathbf N_k$ denote the error transition matrix, noise Jacobian, and discrete noise covariance over the sampling interval $\dt$, respectively, while the nominal state integrates~\eqref{eq:kinematics} with constant biases and zero noise.
The learned displacement corrects the filter state
\begin{equation}
 \begin{aligned}
 \mathbf y&=\hat{\mathbf d}-(\hat\R_\gamma^{-})^T(\hat\p_j^{-}-\hat\p_i^{-})\\
 \bK&=\bP^{-}\bH^T(\bH\bP^{-}\bH^T+\boldsymbol\Sigma)^{-1}\\
 \hat\bX^{+}&=\hat\bX^{-}\boxplus\bK\mathbf y\\
 \bP^{+}&=(\I-\bK\bH)\bP^{-}
 \end{aligned}
 \label{eq:update}
\end{equation}
where $\bX$ comprises $m$ cloned poses and the current navigation state $\bx$ in~\eqref{eq:state}, and $\hat\bX$ is its estimate.
Here $\mathbf y$ is the innovation, $\bK$ is the Kalman gain, and $\bH$ is the displacement Jacobian with respect to the joint error $\delta\bX\in\mathbb R^{6m+15}$.
Here, all poses in the innovation are evaluated before the current update, and $\hat\R_\gamma^{-}$ is the yaw rotation of the pose clone at $t_i$.
The operator $\boxplus$ denotes local addition on the state manifold.
After the update, pose clones up to $t_i$ are marginalized.

\begin{figure}[!t]
\centering
\includegraphics[width=\columnwidth,trim=0 0bp 0 3bp,clip]{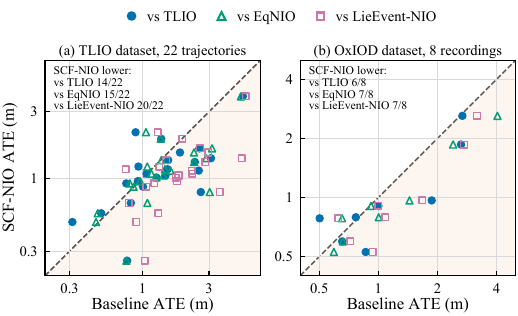}
\caption{Per-sequence ATE of SCF-NIO against TLIO, EqNIO, and LieEvent-NIO (filled circles, open triangles, and open squares) on (a) TLIO and (b) OxIOD, on log axes. Points below the dashed line $y=x$ indicate lower ATE for SCF-NIO. The upper left of each panel gives the count for each baseline. In (a), all 15 points with a baseline ATE above 2.5\,m lie below the diagonal.}
\label{fig:perseq}
\end{figure}

\section{Experiments}
\label{sec:experiments}

We evaluate the trajectory and orientation accuracy of SCF-NIO, and the effect of its shared frame and temporal context.
SCF-NIO is compared with TLIO, EqNIO, and LieEvent-NIO baselines on the TLIO and OxIOD datasets.

\begin{table*}[!t]
\centering
\caption{Trajectory errors in meters on the TLIO and OxIOD datasets. Lower is better. Bold marks the best value in each column.}
\label{tab:main}
\footnotesize
\setlength{\tabcolsep}{4.6pt}
\begin{tabular}{@{}lcccccccccc@{}}
\toprule
 & \multicolumn{5}{c}{TLIO dataset} & \multicolumn{5}{c}{OxIOD dataset} \\
\cmidrule(lr){2-6}\cmidrule(l){7-11}
 & Mean & Median & & Endpoint & Window & Mean & Median & & Endpoint & Window \\
Method & ATE & ATE & RTE & error & RMSE & ATE & ATE & RTE & error & RMSE \\
\midrule
TLIO~\cite{liu2020tlio} & 1.6094 & 1.3101 & 1.3986 & 2.1611 & 0.1311 & 1.3724 & 0.9318 & 1.1714 & 2.0180 & 0.1722 \\
EqNIO~\cite{jayanth2025eqnio} & 1.6345 & 1.3044 & 1.4364 & 2.1725 & 0.1292 & 1.4655 & 0.9615 & 1.2289 & 1.9184 & 0.1847 \\
LieEvent-NIO~\cite{jayanth2025lieevents} & 2.0687 & 1.7514 & 1.8131 & 2.5156 & 0.1385 & 1.4942 & 1.0411 & 1.3113 & 2.2468 & 0.1879 \\
SCF-NIO & \textbf{1.2409} & \textbf{1.1071} & \textbf{1.0016} & \textbf{1.8292} & \textbf{0.0819} & \textbf{1.1319} & \textbf{0.8509} & \textbf{1.1414} & \textbf{1.7409} & \textbf{0.1459} \\
\bottomrule
\end{tabular}
\end{table*}

\begin{figure*}[!t]
\centering
\includegraphics[width=.95\textwidth]{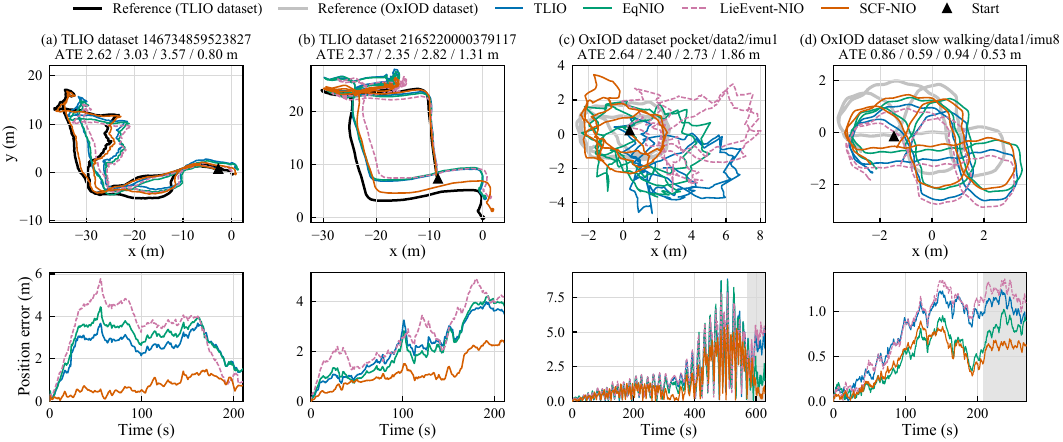}
\caption{Example trajectories in the horizontal plane in the top row and 3D position error over time in the bottom row. (a) and (b) TLIO dataset trajectories 146734859523827 and 2165220000379117, the latter a rectangular loop; (c) OxIOD dataset pocket/data2/imu1; (d) OxIOD dataset slow walking/data1/imu8. The triangle marks the first plotted update. The OxIOD dataset recordings repeat loops in a motion capture room, so their reference is drawn as a wide gray line, and the top panels of (c) and (d) show only the last 60\,s, shaded below, while ATE covers the whole recording. The examples are among the largest relative ATE reductions against TLIO on each dataset.}
\label{fig:traj}
\end{figure*}

\subsection{Datasets and Evaluation Protocol}
\label{subsec:experimental_setup}

The TLIO dataset~\cite{liu2020tlio} is split into 283 training, 35 validation, and 36 test sequences, and trajectories are evaluated on the 22 test sequences with complete raw IMU data.
From the OxIOD dataset~\cite{chen2018oxiod}, 109 recordings are split by recording into 90 training, 11 validation, and 8 test recordings.
All trajectory comparisons use 200\,Hz inertial input, 20\,Hz filter updates, and a five second warmup.
All methods receive the same reference initialization and clone reset at the first update and then run in closed loop on raw IMU data.
The ATE of sequence $s$ compares the estimates $\hat{\mathbf p}_{s,i}$ with the valid reference positions $\mathbf p_{s,i}$ at its $K_s$ scored samples
\begin{equation}
 \operatorname{ATE}_s
 =\sqrt{\frac{1}{K_s}\sum_{i=1}^{K_s}
 \|\hat{\mathbf p}_{s,i}+\mathbf b_s-\mathbf p_{s,i}\|^2}
 \label{eq:ate}
\end{equation}
where $\mathbf b_s$ aligns the first common scored position by translation only.
With the aligned error $\mathbf e_{s,i}=\hat{\mathbf p}_{s,i}+\mathbf b_s-\mathbf p_{s,i}$, the relative trajectory error (RTE) is the root mean square of $\mathbf e_{s,j}-\mathbf e_{s,i}$ over all pairs of scored updates separated by 60\,s, and endpoint error is the norm of $\mathbf e_{s,i}$ at the last scored sample.
Window RMSE is the 3D RMSE of the online one second displacement predictions against the reference displacement.
We compute each error per sequence. Table~\ref{tab:main} reports the mean and median ATE, the mean RTE and endpoint error, and the mean window RMSE for TLIO and median window RMSE for OxIOD. Each sequence has equal weight.
We report the median predicted horizontal standard deviation across all test windows.
Paired comparisons use identical scoring timestamps and reference validity masks and report the number of sequences on which SCF-NIO has the lower error and the relative change of the mean.

\subsection{Training and Compared Methods}
\label{subsec:training_setup}

In SCF-NIO, $\phi$ is a one dimensional ResNet of four stages of two blocks with a $1\times1$ convolution to 128 channels, and the one layer LSTM and both fully connected branches of $\psi$ have 128 hidden units.
SCF-NIO is trained with Adam using a learning rate of $10^{-4}$, a batch size of 128, gradient clipping at 0.1, and mixed precision.
Training minimizes the NLL of~\eqref{eq:nll}, with the uncertainty branch detached for the first nine epochs.
The checkpoint with the lowest validation NLL from epoch 10 onward is selected.
The TLIO, EqNIO, and LieEvent-NIO baselines take the latest one second window, whereas SCF-NIO takes a five second history of $W=5$ windows of $L=200$ samples.
LieEvent-NIO uses its released variant and the filter orientation and initial velocity of each window.
On the TLIO dataset, all methods are trained for 100 epochs, and on the OxIOD dataset every model is trained for 200 epochs.
There, TLIO and SCF-NIO share valid history sampling, batchwise augmentation, and the NLL.
EqNIO and LieEvent-NIO keep their native recipes with an initial MSE training.

\subsection{Accuracy Compared with Baselines}
\label{subsec:main_results}

On the TLIO dataset, SCF-NIO attains the lowest value of every metric, as shown in Table~\ref{tab:main}.
Its mean ATE of 1.2409\,m is 22.9\%, 24.1\%, and 40.0\% below those of TLIO, EqNIO, and LieEvent-NIO, respectively.
RTE falls by 28.4\%, 30.3\%, and 44.8\% on 19, 21, and 22 trajectories, and window RMSE falls by 37.6\%, 36.7\%, and 40.9\% on all 22 trajectories.
Endpoint error falls by 15.4\%, 15.8\%, and 27.3\% on 13, 14, and 15 trajectories.
On the OxIOD recordings, SCF-NIO also has the lowest value in all five columns of Table~\ref{tab:main}.
Its mean ATE is 1.1319\,m, a reduction of 17.5\%, 22.8\%, and 24.2\% relative to TLIO, EqNIO, and LieEvent-NIO, respectively. Its ATE is lower on 6, 7, and 7 of the 8 recordings.
Mean RTE falls by 2.6\%, 7.1\%, and 13.0\%, with lower RTE on 5, 6, and 7 recordings, respectively.
On TLIO, the median predicted horizontal standard deviation is 0.0401\,m for SCF-NIO and 0.0887 to 0.0941\,m for the baselines. On OxIOD, the corresponding values are 0.0398\,m and 0.0833 to 0.0888\,m.
Relative to the baselines, SCF-NIO reduces mean window RMSE by 36.7\% to 40.9\% on TLIO and 6.1\% to 11.1\% on OxIOD recordings.

We split the 22 TLIO dataset trajectories of Fig.~\ref{fig:perseq} at the median ATE of TLIO, 1.3101\,m, into 11 harder and 11 easier ones.
Against TLIO, SCF-NIO has lower ATE on 10 of the harder ones, whose mean falls by 34.4\%, from 2.3760 to 1.5588\,m, and on 4 of the easier ones, whose mean rises by 9.5\%, from 0.8429 to 0.9229\,m.
Fig.~\ref{fig:traj} shows four large reductions, including the largest, from 2.6184 to 0.7964\,m, and the largest increase is from 0.8922 to 2.1377\,m.
Against TLIO on the OxIOD dataset, ATE falls on the running, pocket, and slow walking recordings, for example from 2.6383 to 1.8640\,m on the sequence pocket/data2/imu1.
\begin{figure}[!htb]
\centering
\includegraphics[width=\columnwidth,trim=0 5bp 0 0bp,clip]{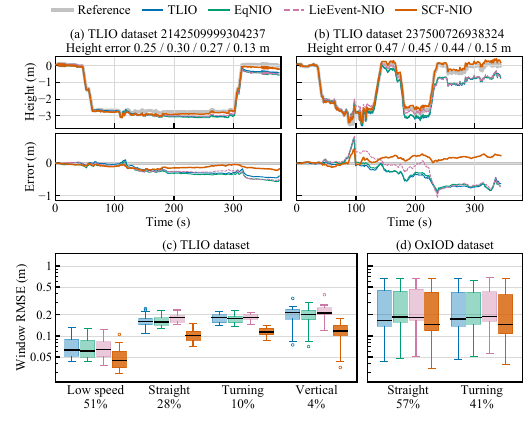}
\caption{Height on trajectories with stairs and window RMSE by motion class. (a) and (b) Two TLIO dataset trajectories, with the height error of TLIO, EqNIO, LieEvent-NIO, and SCF-NIO above and the estimate minus the reference below. (c) and (d) Per sequence window RMSE on log axes, with the share of windows below each class. Boxes span the middle 50\% of sequences, lines mark the medians, whiskers extend at most 1.5 times the RMSE range of each box, and circles mark sequences beyond the whiskers.}
\label{fig:motion}
\end{figure}
Two of the four TLIO dataset trajectories with stairs appear in Fig.~\ref{fig:motion}(a) and (b).
Over these four, the height error, the ATE of the vertical component, falls by 42.6\%, 49.9\%, and 56.0\% on 3, 3, and 4 of them.
Fig.~\ref{fig:motion}(c) and (d) group the windows of each sequence by reference motion.
\begin{figure}[!h]
\centering
\includegraphics[width=\columnwidth,trim=0 4bp 0 3bp,clip]{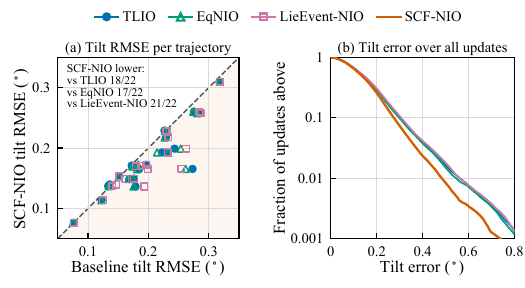}
\caption{Tilt error of the filter on the 22 TLIO dataset trajectories. (a) Per trajectory tilt RMSE of SCF-NIO against TLIO, EqNIO, and LieEvent-NIO. Points below the dashed line are trajectories on which SCF-NIO has the lower error. (b) Fraction of the updates of all trajectories whose tilt error exceeds the value on the horizontal axis, on a log axis.}
\label{fig:orientation}
\end{figure}
Low speed denotes a horizontal speed below 0.2\,m/s, a heading change of $30^\circ$ per second separates straight walking from turning, and vertical motion denotes a vertical speed above 0.2\,m/s.
On the TLIO dataset, SCF-NIO lowers window RMSE on every trajectory of each class, by 26.5\% to 30.3\% at low speed, by 34.8\% to 42.1\% during straight walking and turning, and by 39.1\% to 47.0\% during vertical motion.
On OxIOD, straight walking and turning account for 98.0\% of the windows. Across the three baselines, SCF-NIO reduces the median window RMSE over recordings by 13.3\% to 22.1\% during straight walking and by 17.0\% to 23.3\% during turns.

The filter also estimates orientation, which Fig.~\ref{fig:orientation} evaluates on the TLIO dataset.
At every update, the tilt error is the angle between the estimated and reference gravity directions in the body frame, and the yaw error is the heading difference.
Both are computed without alignment, because all methods start from the reference state.
The mean tilt RMSE of SCF-NIO is 0.181$^\circ$, which is 9.6\%, 8.9\%, and 11.5\% below those of TLIO, EqNIO, and LieEvent-NIO, and it is lower on 18, 17, and 21 trajectories.
The reduction is largest in the tail of Fig.~\ref{fig:orientation}(b), where 0.83\% of the updates exceed 0.5$^\circ$ against 1.50\% to 1.83\% for the baselines.
Roll and pitch are observable through the coupling of gravity and velocity, so the more accurate displacement measurement of SCF-NIO also lowers the tilt error.
SCF-NIO also attains the lowest mean yaw RMSE, 1.88$^\circ$ against 1.94$^\circ$ to 1.99$^\circ$, with a smaller margin because global yaw is unobservable from~\eqref{eq:displacement} and drifts as the gyroscope errors integrate.
\begin{figure}[!t]
\centering
\includegraphics[width=\columnwidth,trim=0 3bp 0 0bp,clip]{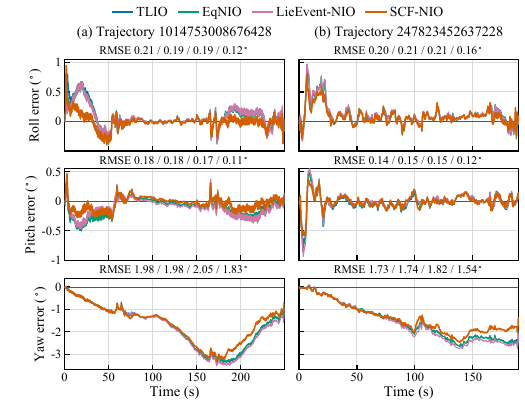}
\caption{Euler angle errors of TLIO, EqNIO, LieEvent-NIO, and SCF-NIO on the TLIO dataset trajectories (a) 1014753008676428 and (b) 247823452637228. Rows show roll, pitch, and yaw in the ZYX convention, and the RMSE above each panel lists the four methods in this order.}
\label{fig:euler}
\end{figure}

Fig.~\ref{fig:euler} shows the Euler angle errors on the two trajectories with the largest reductions, on which SCF-NIO lowers the roll, pitch, and yaw RMSE against every baseline.
On trajectory 1014753008676428, the roll and pitch RMSE fall to 0.12$^\circ$ and 0.11$^\circ$ from at least 0.19$^\circ$ and 0.17$^\circ$, because SCF-NIO removes most of the roll and pitch offsets that the baselines keep during the first 50\,s and after 170\,s.
At the end of the two trajectories, the yaw error of SCF-NIO is 0.96$^\circ$ and 1.71$^\circ$, against 1.06$^\circ$ to 1.22$^\circ$ and 2.32$^\circ$ to 2.45$^\circ$ for the baselines.

% \FloatBarrier
\subsection{Ablation Study}
\label{subsec:ablation}

Ablations on the TLIO dataset test the temporal prior in a shared frame, as shown in Table~\ref{tab:ablation}.
Each trained variant changes one module of SCF-NIO and keeps its data, seed, loss, and number of epochs.
Without canonical frame, the predictor takes the gravity aligned history.
Without shared frame, each window has its own frame, and the frame of the latest window maps the output back.
Without temporal context, the predictor takes only the latest window.
\begin{table}[!t]
\centering
\caption{Ablation on the TLIO trajectories. Each variant removes one module of SCF-NIO. Errors are in meters. Lower is better.}
\label{tab:ablation}
\footnotesize
\setlength{\tabcolsep}{4pt}
\begin{tabular}{@{}lccc@{}}
\toprule
Variant & Mean ATE & RTE & Window RMSE \\
\midrule
SCF-NIO & 1.2409 & 1.0016 & 0.0819 \\
\midrule
Without canonical frame & 1.3058 & 1.0801 & 0.0839 \\
Without shared frame & 1.3859 & 1.1221 & 0.1013 \\
Without temporal context & 1.4168 & 1.1771 & 0.1192 \\
\bottomrule
\end{tabular}
\end{table}

Against these variants, SCF-NIO lowers window RMSE by 2.4\%, 19.2\%, and 31.3\%, RTE by 7.3\%, 10.7\%, and 14.9\%, and mean ATE by 5.0\%, 10.5\%, and 12.4\%, respectively.
The variant without shared frame also has higher window RMSE than the variant without canonical frame on all 22 trajectories, 0.1013 against 0.0839\,m, and higher mean ATE, 1.3859 against 1.3058\,m.
Against the three variants, SCF-NIO also lowers window RMSE in every motion class defined in Section~\ref{subsec:main_results}, and it lowers height error on the four trajectories with stairs by 36.2\%, 23.8\%, and 18.8\%, with lower height error on 4, 4, and 3 of them.

SCF-NIO thus attains the lowest value of every metric in Table~\ref{tab:ablation}, and each module of the predictor lowers all three errors.
The temporal context gives the largest reductions, and the canonical frame helps only when all windows share it.

% \FloatBarrier
\begin{table}[!h]
\centering
\caption{Computational cost of four methods.}
\label{tab:runtime}
\footnotesize
\setlength{\tabcolsep}{4pt}
\begin{tabular}{@{}lccccc@{}}
\toprule
 & & Params & MACs & Latency & Filter time \\
Model & Input & (M) & (M) & (ms) & (ms) \\
\midrule
TLIO & 1\,s & 5.42 & 39 & 1.7 & 186 \\
EqNIO & 1\,s & 6.02 & 206 & 3.6 & 218 \\
LieEvent-NIO & 1\,s & 5.43 & 39 & 1.7 & 396 \\
SCF-NIO & 5\,s & 5.07 & 356 & 4.2 & 272 \\
\bottomrule
\end{tabular}
\end{table}

\subsection{Computational Cost and Limitations}
\label{subsec:limitations}

Table~\ref{tab:runtime} compares the computational cost of the four methods.
The networks run on an NVIDIA GeForce RTX 2080 Ti GPU, and the ESKF uses one thread of an Intel Xeon Platinum 8255C CPU.
MACs count multiply accumulate operations in convolutions, linear layers, and the LSTM.
Latency is averaged over 1000 TorchScript calls, and filter time measures processing time per second of IMU data.
SCF-NIO predicts a displacement in 4.2\,ms on average and processes one second of IMU data through the complete filter in 272\,ms.
This supports real time operation at the 20\,Hz update rate.

Compared with EqNIO, SCF-NIO reads five times as much IMU history with 1.7 times the MACs, 16\% fewer parameters, and 0.6\,ms more latency.
When timed separately, the EqNIO frame module takes 1.7\,ms for one window. The SCF-NIO frame module takes 2.1\,ms for five windows, an increase of 0.4\,ms.
SCF-NIO predicts the shared frame from the latest window and applies it to all $WL$ samples. The CNN encodes each window, and the LSTM processes the resulting sequence of window features.
One update therefore costs $\mathcal O(WL)$ operations, against $\mathcal O(L)$ for TLIO and EqNIO.
Because the shared frame changes at every update, each update encodes all $W$ windows anew.
Each added window costs 37.8\,M MACs, but the GPU encodes the windows in one batch, so the latency of SCF-NIO stays between 4.2 and 4.3\,ms.

% \FloatBarrier

\section{Conclusion}
\label{sec:conclusion}

This letter presents SCF-NIO, a neural inertial odometry method.
It constructs a canonical frame shared by inertial measurements across consecutive windows and learns a temporal displacement prior in this frame.
SCF-NIO outperforms three neural inertial baselines in mean ATE, with reductions of 22.9\% on the TLIO dataset and 17.5\% on OxIOD compared with the TLIO baseline. It also achieves the lowest window RMSE on all 22 TLIO test sequences.
Future work will fuse this temporal prior into VIO to constrain drift when visual tracking fails.

\begingroup
\footnotesize
\bibliographystyle{IEEEtranN}
\bibliography{customabrv,IEEEabrv,paper}
\endgroup

\end{document}